\documentclass[
]{ceurart}

\usepackage{graphicx}
\usepackage{booktabs}
\usepackage{amsmath}
\begin{document}

\copyrightyear{2026}
\copyrightclause{Copyright for this paper by its authors.
  Use permitted under Creative Commons License Attribution 4.0
  International (CC BY 4.0).}

\conference{}

\title{LLM-Assisted Discovery of Typed Semantic Links for Ontology Network Construction}

\author[1]{Nouha Hayouni}[%
email=nouha.hayouni@uni-passau.de,
]
\address[1]{Chair of Data and Knowledge Engineering, University of Passau, Germany}

\author[2]{Sheeba Samuel}[%
orcid=0000-0002-7981-8504,
email=sheeba.samuel@informatik.tu-chemnitz.de,
]
\cormark[1]
\address[2]{Distributed and Self-organizing Systems, Chemnitz University of Technology, Germany}

\author{Alsayed Algergawy}[%
orcid=0000-0002-8550-4720,
email=alsayed.algergawy@uni-jena.de,
]

\cortext[1]{Corresponding author.}

\begin{abstract}
Constructing typed, justified semantic links between ontologies is essential for enabling interoperability across heterogeneous and interdisciplinary knowledge domains. However, manually curating such links is difficult to scale. To address this challenge, we propose an end-to-end framework for ontology network construction that automates the discovery and generation of both intra-domain and inter-domain relationships. Our approach combines domain-adapted DistilBERT embeddings for dense contextual representation, clustering-based pre-filtering to reduce the candidate search space, and GPT-4o-driven relationship generation via iterative prompt engineering to produce semantically rich, interpretable links.
Applied to ReproduceMeON—a network of 33 ontologies spanning machine learning, microscopy, computational science, and experimental workflow—the pipeline reduces approximately 800k raw concept pairs to 95k high-quality candidates.
Human expert validation of 429 generated relationships by two independent annotators yields an overall precision of 80.19\% (91.49\% on high-certainty annotations) and an F1 of 0.890, with substantial inter-annotator agreement.
Comparative experiments against five similarity-based baselines, including Sentence-BERT, show a substantial performance gap (best baseline F1 = 0.581), while an ablation study demonstrates that similarity-based methods alone fail to discriminate valid from invalid relationships (AUC\,$\approx$\,0.5) on the filtered candidate set. These findings highlight the necessity of LLM-based reasoning over concept roles and domain semantics for accurate relationship construction. 
\end{abstract}


\maketitle

\section{Introduction}
\label{sec:intro}
In modern scientific environments, interdisciplinary studies often span multiple knowledge domains—ranging from experimental protocols to machine learning (ML) algorithms and computational workflows. Attempting to unify all relevant knowledge within a single monolithic ontology is not only impractical but also counterproductive. Instead, ontology networks (ONs), which interlink multiple ontologies through typed semantic relationships, provide the formal backbone for such cross-domain knowledge integration~\cite{haased2006Network,samuel2023the}. The development of such networks relies heavily on the accurate identification of both intra-domain relationships—links between concepts from different ontologies within the same domain as well as inter-domain relationships—links connecting concepts across distinct domains. 

Existing approaches fall short. Classical alignment tools (LogMap~\cite{2011logmap}, AML ~\cite{2013agreementmakerlight}) identify equivalent concepts but do not generate typed, justified relationships~\cite{samuel2023the}. Embedding-based methods such as OWL2Vec*~\cite{chen2021owl2vec} and RDF2Vec~\cite{ristoski2016rdf2vec} improve scalability but cannot discriminate between valid and invalid relationships among high-similarity candidates: as our ablation study confirms empirically, all five similarity methods—including Sentence-BERT—achieve AUC\,$\approx$\,0.5 on the pre-filtered candidate set, demonstrating that semantic similarity alone is insufficient. LLM-based approaches~\cite{kommineni2024human} have explored GPT-driven ontology enrichment but without systematic pre-filtering, iterative prompt engineering, or role-based materialization.

We present a complete end-to-end pipeline that combines domain-adapted DistilBERT~\cite{26} fine-tuning, clustering-based 90th-percentile candidate filtering, and GPT-4o~\cite{31}-driven relationship generation via structured prompt engineering. Applied to ReproduceMeON, the pipeline reduces 800k raw pairs to 95k high-quality candidates and produces typed, justified semantic links with provenance. Human expert validation of 429 generated relationships by two independent annotators yields a precision of 80.19\% overall (91.49\% on high-certainty annotations), with substantial inter-annotator agreement (Cohen's $\kappa = 0.723$, 90.5\% on 274 dually annotated pairs). On a balanced 929-pair evaluation set, our pipeline achieves F1\,=\,0.890 and AUC-ROC\,=\,0.927, outperforming the best similarity-based baseline (Sentence-BERT, F1\,=\,0.581) by 30.9 percentage points. 
All code, data, and the expert-annotated validation set are available in GitHub: \url{https://github.com/fusion-jena/ReproduceMeON/tree/main/LLM_based_link_discovery}. 
\par To sum up, the main contributions of this paper are:
\begin{enumerate}
    \item \textbf{End-to-end reproducible pipeline}: A complete, open-source pipeline combining DistilBERT fine-tuning, 90th-percentile clustering-based candidate selection (800k~$\rightarrow$~95k pairs), iterative GPT-4o prompt engineering, and role-based mapping for automated ON development.
    \item \textbf{Domain-aware prompt engineering}: A structured, iterative prompt refinement methodology—including contextual, task-specific, chain-of-thought, and few-shot strategies—that steers GPT-4o towards precise, domain-aligned relationship generation with scientific justifications.
    \item \textbf{Role-based link mapping}: A mapping step that converts LLM-generated typed links into workflow-oriented semantic roles (e.g., Provider, Consumer, Enabler) grounded in established Ontology Design Patterns.
    \item \textbf{Empirical evaluation with quantitative baselines}: Dual expert validation of 429 relationships (274 with measured inter-annotator agreement, Cohen's $\kappa = 0.723$, \textit{substantial agreement}) achieving precision 80.19\% (91.49\% high-certainty), F1\,=\,0.890, AUC-ROC\,=\,0.927—outperforming the best similarity baseline by 30.9 pp in F1—with an ablation study confirming that LLM reasoning, not similarity, drives the precision gain.
\end{enumerate}

Section~\ref{sec:relatedwork} reviews related work. Section~\ref{sec:method} details the pipeline. Section\ref{sec:results} presents and analyses the experimental results. Section\ref{sec:conclusion} concludes and outlines future directions.

\section{Related Work}
\label{sec:relatedwork}

Ontology networks (ONs) have emerged as a solution to address the limitations of isolated ontologies by enabling semantic interoperability and knowledge sharing across domains \cite{38,37,1}. Traditional approaches to ON construction have relied heavily on manual curation or rule-based alignment techniques, which are labor-intensive, domain-specific, and often struggle with scalability and adaptability \cite{38,37}. These approaches typically involve schema matching, logical alignment, and ontology mediation strategies using description logic, heuristics, or statistical methods \cite{samuel2023the,38,37,14,reyes2021ontology}. While effective in controlled settings, such techniques are insufficient for large-scale, dynamic, or interdisciplinary knowledge integration.

More recently, machine learning and embedding-based methods have been introduced to support ontology alignment and concept matching. Techniques such as RDF2Vec \cite{ristoski2016rdf2vec} and OWL2Vec* \cite{chen2021owl2vec} use structural and contextual features of ontologies to generate embeddings that enable similarity-based reasoning. These models have improved the scalability of ontology alignment but are often limited in capturing deeper semantic nuances, especially when relationships require contextual or domain-specific interpretation.

The advent of LLMs, such as BERT \cite{18} and GPT \cite{17}, has opened new avenues for semantic understanding in natural language and knowledge graphs. These models excel at extracting and reasoning about relationships between concepts across unstructured and semi-structured data. Several recent studies have explored the use of LLMs for ontology development and enrichment \cite{kommineni2024human}, relation extraction \cite{15}, and knowledge graph construction \cite{kommineni2024towards}. However, their integration into the ON development process—particularly for generating both intra- and inter-domain links in structured, role-aware formats—remains underexplored.

Prompt engineering has further emerged as a powerful strategy for controlling the behavior of LLMs in specialized tasks. 
Ambiguously phrased prompts might lead to vague answers or hallucinations \cite{21}. Nonetheless, the application of prompt engineering for generating ontology relationships that are both semantically meaningful and structurally usable within an ON is still in its early stages.

Classical alignment tools such as LogMap and AML~\cite{samuel2023the} achieve high precision on equivalence mappings but do not generate typed, justified relationships or cover the full semantic spectrum (functional, causal, temporal, etc.). Embedding-based methods like OWL2Vec*~\cite{chen2021owl2vec} learn structural representations but still require a separate classifier for relationship typing. LLM-based approaches~\cite{kommineni2024human} leverage GPT models for ontology tasks but without the pre-filtering, role-based materialization, and iterative prompt refinement that our pipeline provides. Our method uniquely combines all four aspects in a single, end-to-end framework validated on a real multi-domain ontology network. A quantitative comparison of similarity-based baselines against our pipeline (Precision/Recall/F1) is provided in Section~\ref{sec:results}.

Our task is \textit{typed semantic link generation for ontology networks}: producing new, typed, semantically grounded associations between concept classes across ontologies that do not yet have explicit links, with the goal of enriching the ON's relational structure. The generated links represent expert-validated typed semantic associations (functional, hierarchical, causal, etc.) rather than formal OWL axioms with logical guarantees—no satisfiability checking, domain/range validation, or reasoner-based consistency checking is performed. This scope distinction explains why classical alignment systems (AML, LogMap) and embedding-based matchers are not directly comparable: they find equivalences, not functional or causal associations between non-equivalent concepts.

Our work contributes a systematic, reproducible integration of four components—domain-adapted embeddings, clustering-based pre-filtering, structured LLM prompting, and role-based materialization—into a single validated pipeline for ON link generation. While each component builds on established techniques, their principled combination for the specific challenge of typed, justified inter-ontology link discovery at the scale of a multi-domain ON (33 ontologies, 800k candidate pairs) has not been demonstrated in prior work.

\section{Methodology}
\label{sec:method}
Given an ontology network $\mathcal{N} = \{O_1, \ldots, O_n\}$ where each $O_i$ is an OWL ontology with a set of classes $C_i$, the task is to discover a set of typed, justified links $\mathcal{L} = \{(c_a, r, c_b, j) \mid c_a \in C_i,\, c_b \in C_j,\, i \neq j \text{ or same domain},\, r \in \mathcal{R},\, j \in \text{text}\}$, where $\mathcal{R}$ is a predefined set of typed semantic relationship categories and $j$ is a natural-language justification. A link $(c_a, r, c_b)$ is considered \textit{valid} if a domain expert confirms that the stated relationship $r$ holds between $c_a$ and $c_b$ in the context of their respective ontology domains. The task explicitly excludes equivalence mappings (handled by classical alignment tools) and focuses on \emph{functional, causal, compositional, and instrumental} relationships that enrich the ON's cross-domain expressivity.

\begin{figure}
    \centering
    \includegraphics[width=1.0\textwidth]{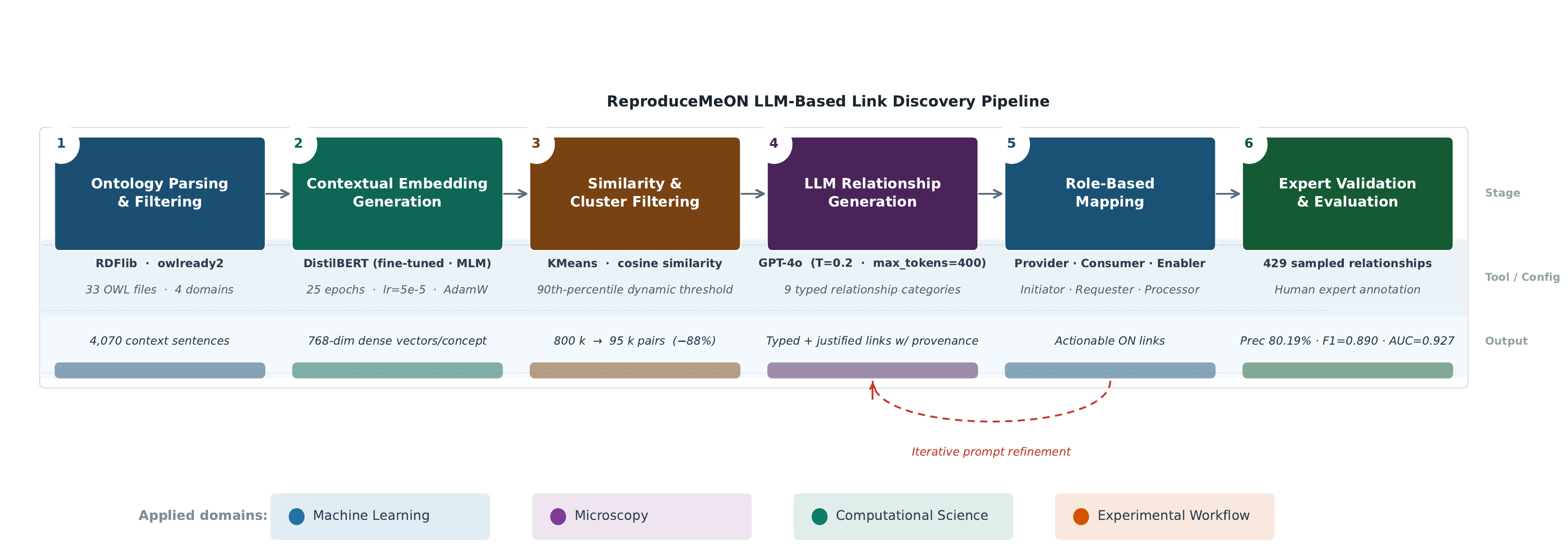}
    \caption{Overview of the proposed end-to-end pipeline for discovering and generating intra- and inter-domain links for the development of ReproduceMeON ontology network.}
    \label{fig:pipeline}
\end{figure}

We propose a six-stage pipeline (Fig.~\ref{fig:pipeline}) applied to ReproduceMeON~\cite{samuel2023the,1}: (1)~\textit{Ontology Parsing \& Filtering}—extract class-type RDF triples; (2)~\textit{Contextual Embedding Generation}—fine-tune DistilBERT~\cite{26} on domain-enriched sentences; (3)~\textit{Similarity \& Cluster Filtering}—90th-percentile cosine threshold reduces 800k to 95k pairs; (4)~\textit{LLM Relationship Generation}—GPT-4o~\cite{31} assigns typed, justified links; (5)~\textit{Role-Based Mapping}—converts links to actionable semantic roles; (6)~\textit{Expert Validation \& Evaluation}—human annotation and quantitative baselines.

\subsection{Ontology Parsing \& Filtering}
OWL ontologies are parsed with owlready2\footnote{\url{https://owlready2.readthedocs.io/en/v0.47/}} from the ReproduceMeON repository\footnote{\url{https://github.com/fusion-jena/ReproduceMeON}} to extract subject–predicate–object triples; a filtering step retains only class-type triples (e.g., \texttt{DataProcessingAlgorithm subClassOf Algorithm}), removing literals and non-class data. Additional context—rdfs:label, rdfs:comment, and annotations—is extracted to disambiguate domain-specific meanings (e.g., ``process'' in ML vs.\ experimental settings) and enrich subsequent embeddings.

\subsection{Contextual Embedding Generation}
To adapt DistilBERT \cite{26} to domain-specific knowledge, RDF triples are converted into context sentences  (e.g., `WebServiceAlgorithm rdfs:subClassOf DataProcessingAlgorithm") enriched with corresponding \textit{rdfs:label} and \textit{rdfs:comment} metadata. This contextualization helps the model learn subtle semantic distinctions between concepts. The fine-tuning dataset includes subject, object, and descriptive fields, split into 80\% training, 10\% validation, and 10\% test sets. After iterative tuning, the final configuration uses a batch size of 16, a learning rate of 5e-5, and 25 epochs. Training employs the Masked Language Modeling (MLM) objective, randomly masking 15\% of tokens in each sentence to encourage contextual prediction, thereby enhancing the model’s capacity to capture subject-object semantics.

To prevent overfitting and ensure generalization, we use: (i) AdamW Optimizer to decouple weight decay from gradient updates \cite{loshchilov2019decoupled}, (ii) Early Stopping if validation loss stagnates for 3 epochs, (iii) Validation Set Checks using 10\% of data, and (iv) Best Model Checkpointing to retain the model with the lowest validation loss.

Input sentences are tokenized using WordPiece (256 tokens, padding/truncation)~\cite{18} and processed through the fine-tuned model; mean pooling over hidden states yields \textit{Sentence Embeddings} (full subject–object context) and \textit{Subject/Object Embeddings} (hierarchical and contextual representations, e.g., ``Classification Algorithm'' as a subclass of ``Algorithm''). Embeddings are generated in batches with parallel processing. KMeans clustering (cluster count selected via Elbow and Silhouette methods)~\cite{32} organizes embeddings into semantically coherent groups for downstream relationship identification.

\subsection{Similarity \& Cluster Filtering}  
This step aims to compute semantic similarity scores between concepts based on their embeddings, using cluster membership to limit comparisons and ensure meaningful, efficient pair generation. Cosine similarity is calculated only within each cluster, focusing on semantically relevant ties and reducing computational overhead. Using scikit-learn’s cosine similarity function, pairwise similarities are computed within clusters.

\textbf{Threshold selection.} The similarity threshold directly controls the precision–recall trade-off for candidate pair generation. We adopt a \textit{dynamic threshold} set at the 90th percentile of within-cluster cosine similarity scores. This data-driven choice adapts to each cluster’s score distribution rather than imposing a single global cutoff: dense clusters (e.g., Cluster 1, which contained intra-domain microscopy concepts) produce tighter distributions and thus higher thresholds, while sparser clusters produce lower thresholds that still filter out noise. As a sensitivity check, a fixed threshold of 0.97 was also evaluated; it produced 30\% fewer candidate pairs with no measurable gain in downstream validation precision, confirming that the percentile-based approach retains more semantically useful candidates. For each valid pair, metadata including subject and object classes, domains, ontology sources, and similarity scores is recorded. This structured approach ensures that generated pairs reflect semantically coherent relationships within their cluster context. To maintain uniqueness, duplicate pairs are removed in a final data cleaning step.

\subsection{LLM Relationships Generation}
This step is central to the development of ReproduceMeON, automating the generation of semantic relationships between ontology concepts based on similarity scores and contextual understanding. Each concept pair—with its ontology sources, domains, and similarity score—is submitted to GPT-4o to determine whether a meaningful semantic relationship exists, considering not only cosine similarity but also concept roles, dependencies, and real-world interactions. The output is a structured dataset of typed, justified links with provenance.
To ensure consistency, the model is guided by nine predefined relationship types inspired by the state of the art~\cite{guarino2009what,ian2001towards}:
\begin{itemize}
\item \textbf{Hierarchical} (parent–child, subtype): capture generalization and specialization relations 
, supporting taxonomy construction, inheritance, and logical reasoning.
\item \textbf{Functional} (enables, requires): describe dependencies and operational flows 
, enabling the modeling of workflows and reusable process components.

\item \textbf{Causal} (cause–effect): represent directional influence between entities, supporting explanation, hypothesis testing, and predictive reasoning.

\item \textbf{Temporal} (sequence-based): encode ordering and timing constraints 
, essential for workflows, protocols, and simulations.

\item \textbf{Spatial} (proximity, location): provide context on physical or spatial configurations, particularly relevant in domains such as microscopy, biology, and robotics.

\item \textbf{Compositional} (part–whole): model structural relationships between components, enabling reasoning over modular and nested systems.

\item \textbf{Comparative} (relative differences): express distinctions or evaluations, supporting decision-making, benchmarking, and recommendation scenarios.

\item \textbf{Transformational} (conversion, evolution): capture state or format changes, important for representing data pipelines, chemical processes, or learning stages.

\item \textbf{Instrumental} (tool–method): identify tools or agents used to perform tasks, supporting the representation of operational roles in computational and experimental workflows.
\end{itemize}
Together these cover the full semantic spectrum needed for interdisciplinary ontology network construction—from structural taxonomies to operational workflows and cross-domain dependencies.
Using the OpenAI API, 
input data and instructions are sent to GPT-4o.
The output includes the relationship type, source and target concepts, a justification sentence, and any cited reference. 
The result is a structured dataset of domain-specific, justified semantic links, enabling fine-grained mapping between ontology concepts and supporting expert validation through transparent reasoning and evidence.

\subsubsection{Prompt Structure and Example}

Each concept pair is submitted via a structured prompt (see example in Fig~\ref{fig:prompt_example}): the \textit{system} message establishes GPT-4o's role as an ontology expert, prohibits vague relations, and mandates a scientific justification; the \textit{user} message supplies both concepts, their ontology sources, domains, and cosine similarity, then specifies a JSON output schema (relationship type, source, target, justification, citation). 
\begin{figure}[h!tbp]
\centering
\begin{minipage}{0.95\linewidth}
\small
\textbf{System:} \ttfamily{You are an expert in ontology construction and knowledge graph generation. Your task is to construct a network of meaningful relationships between concepts in the ReproduceMeON context. Use your semantic understanding and domain knowledge. Follow these rules strictly: Avoid vague relationships like ``related to''; ensure relationships reflect meaningful real-world dependencies; provide a justification sentence; cite a specific publication to support the justification}.\\[4pt]
\textbf{User:} \textit{Input Analysis}~--- Subject\,1: Activity (Microscopy, omeroriken.owl); Subject\,2: Objective (Computational, dockeronto.owl); Similarity Score: 0.87\\
\textit{Instructions}~--- Define: Relationship (e.g., ``enables'', ``requires''), Relationship Nature (Hierarchical / Functional / Causal / Temporal / Spatial / Compositional / Comparative / Transformational / Instrumental), Source, Target, Justification Sentence, Scientific Reference. Return result in the specified JSON format. If no relationship exists, state ``No Relationship'' and explain why.
\end{minipage}
\caption{Representative prompt used for GPT-4o-based relationship generation.}
\label{fig:prompt_example}
\end{figure}

\subsubsection{Prompt Refinement}
Initial prompts produced vague outputs (e.g., ``is related to’’); we therefore applied iterative prompt engineering combining contextual cues (domain, similarity score), task-specific constraints (allowable types, JSON schema), chain-of-thought decomposition~\cite{23}, and few-shot examples, with feedback from prior outputs incorporated each iteration to reduce vagueness and improve diversity~\cite{21,35}.

Although GPT-4o is used for its instruction-following and structured-output capabilities, the pipeline is model-agnostic: instruction-tuned open-source models (LLaMA~\cite{llama3}, Mistral) could substitute at lower cost, while models with extended context windows could reduce API overhead by batching more pairs per call. Comparing model families is planned future work; current results serve as a GPT-4o baseline.

\subsection{Role-Based Mapping}
Role-based mapping converts abstract LLM-generated links into operational semantic roles—Enabler, Processor, Provider, Consumer, Initiator, Requester—adding actionable meaning beyond the raw relationship type. This design is grounded in established Ontology Design Patterns (ODPs)~\cite{gangemi2005ontology}: the role taxonomy mirrors the \textit{AgentRole} and \textit{Participation} patterns, where entities participate in processes under specific functional capacities. For example, mapping \textit{DataSet} as a Provider and \textit{Algorithm} as a Consumer in ``DataSet Provides Data for Algorithm'' directly instantiates the ODP \textit{Provenance} pattern and makes the link usable in automated data pipeline assembly. Across ReproduceMeON's multi-domain scope, this step ensures that inter-domain links such as ``Experimental Protocol Requires Computational Simulation'' become SPARQL-queryable. Evaluating role assignment correctness via competency questions is planned as future work; the present contribution establishes the mapping methodology and taxonomy.

\section{Experimental Evaluation}
\label{sec:results}
In this section, we address our main research question: How can large language models and contextual embeddings be leveraged to automatically generate accurate and semantically meaningful relationships for ontology network development? 
In the following, we present the experimental setup, datasets, and results.

\subsection{Experimental Setup}
Experiments run on Google Colab (NVIDIA T4 GPU). The pipeline uses the Hugging Face Transformers library for DistilBERT fine-tuning~\cite{26}, PyTorch for training and embedding generation, Scikit-learn for cosine similarity and KMeans clustering, RDFlib for ontology parsing, and the OpenAI GPT-4o API~\cite{31} for relationship generation.

\subsection{Datasets} The datasets used in the evaluation consist of ontologies from four primary domains\footnote{\url{https://github.com/fusion-jena/ReproduceMeON}} (computational, experimental workflow, machine learning, and microscopy) \cite{1} and alignment between ontologies \cite{samuel2023the}. Although the ReproduceMeON network also includes provenance ontologies, provenance concepts are interwoven within the experimental workflow domain and are captured through its triples; a dedicated provenance domain partition was therefore not extracted separately.
The datasets span multiple domains, showcasing the diversity of reproducibility-related concepts, but also introducing challenges related to overlapping semantics and domain-specific terminologies.
\par Ontologies from different domains are parsed and RDF triples are extracted from the ReproduceMeON. The extraction process resulted in a total of 46244 unique subjects, 71426 unique objects, and 214 unique predicates.  
The domain of each ontology is used to classify RDFs into several areas. Table \ref{tab:domain_tuple_count} presents the distribution of RDF triples across the four domains utilized in this study.

\begin{table}[ht]
    \centering
    \small 
    \setlength{\tabcolsep}{3pt} 

    \begin{minipage}[t]{0.32\textwidth} 
        \centering
        \caption{Overview of Tuple Counts by Domain} 
        \label{tab:domain_tuple_count}
        \begin{tabular}{|l|r|}
            \hline
            \textbf{Domain} & \textbf{Count} \\ \hline
            Computational   & 191,919        \\ \hline
            Microscopy      & 49,246         \\ \hline
            ML              & 18,504         \\ \hline
            Experimental    & 12,238         \\ \hline
        \end{tabular}
    \end{minipage}
    \hfill 
    \begin{minipage}[t]{0.63\textwidth} 
        \centering
        \caption{Relationship distribution from sampled pairs} 
        \label{tab:distribution_relationships}
        \begin{tabular}{lr | lr}
            \hline
            \textbf{Rel.} & \textbf{\%} & \textbf{Rel.} & \textbf{\%} \\ \hline
            Enables      & 22.84\% & Contains     & 2.10\%  \\
            Requires     & 13.75\% & Implements   & 1.86\%  \\
            Influences   & 5.83\%  & Represents   & 1.86\%  \\
            Is used in   & 5.59\%  & Detects      & 1.63\%  \\
            Is a type of & 4.66\%  & Describes    & 1.63\%  \\
            Utilizes     & 4.43\%  & Operated     & 1.40\%  \\
            Generates    & 3.50\%  & Config.      & 1.17\%  \\
            Defines      & 2.80\%  & \textit{Rare}\footnote{20+ individual relation types each contributing $<2\%$, including \textit{Processes}, \textit{Applies}, \textit{Extends}, \textit{Controls}, \textit{Measures}, \textit{Depends On}, \textit{Precedes}, and others.} & \textit{33.95\%} \\
            Is used by   & 2.56\%  &              &         \\ \hline
        \end{tabular}
    \end{minipage}
\end{table}

\subsection{Results}
After applying the pipeline on the available datasets to determine intra- and inter-links between concepts from different ontologies, we get an output file containing comprehensive and detailed information.
The figure shows that the model is able to discover an inter-link between the source concept \textit{``Activity"} from the \textit{Microscopy} domain and the target concept \textit{``Objective"} from the \textit{Computational} domain. Furthermore, the model is able to predict and label the inter-link as \textit{``enables"} with the \textit{``Functional"} type. 
Table~\ref{tab:distribution_relationships} shows the distribution of relationship types across all generated links.

\begin{figure}[h!tbp] 
    \centering
    \includegraphics[width=0.7\textwidth,keepaspectratio]{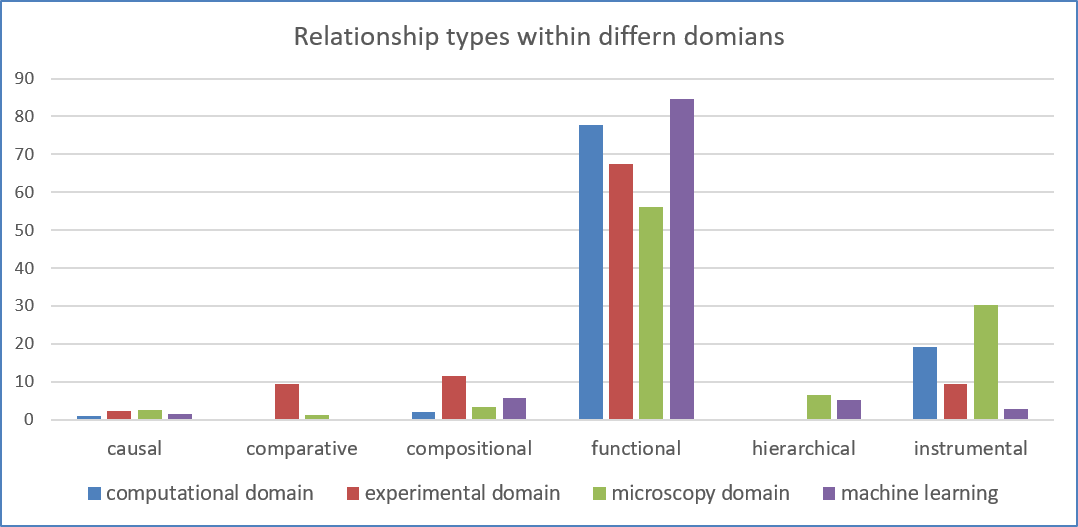}
    \caption{Relation types across different domains}
    \label{fig:relstions_type}
\end{figure}

\subsubsection{Diversity and inter/intra-domain balance}
Table~\ref{tab:distribution_relationships} shows that Enables (22.84\%) and Requires (13.75\%) dominate, reflecting the dependency-heavy nature of scientific workflows, while rarer types (Implements, Configures, Describes) capture domain-specific interactions. Functional relationships dominate across all four domains (ML 84.67\%, Computational 77.66\%, Experimental Workflow 67.44\%, Microscopy 56.13\%), with Instrumental links prominent in Microscopy (30.32\%) and Computational (19.15\%). The Experimental Workflow domain shows the broadest variety, with compositional (11.63\%), comparative (9.30\%), and hierarchical (9.30\%) links also contributing.
The pipeline produces 51.3\% inter-domain and 48.7\% intra-domain links, reflecting a balanced dual emphasis suited to both specialized domain reasoning and cross-disciplinary knowledge reuse.

\subsection{Quantitative Baseline Comparison}
\label{sec:baselines}
To contextualise our results, we compare the LLM pipeline against five similarity-based baselines. The comparison is specifically designed to test the claim that \emph{semantic similarity alone is insufficient for relationship validity judgement}—it is not a claim about outperforming full ontology alignment systems such as AML or LogMap, which produce equivalence mappings rather than typed, justified relationships and therefore address a different task. A comparison with other LLM-based relationship generators is left for future work, as no publicly available system produces the same output format (typed, justified, role-mapped ON links).

The evaluation set comprises (i) the 344 valid annotated relationships (positives), (ii) the 85 invalid annotated relationships, and (iii) 500 randomly generated cross-domain concept pairs as easy negatives. We acknowledge that randomly sampled negatives are easier than real-world hard negatives and thus inflate absolute F1 scores; the F1\,=\,0.890 figure should therefore be interpreted as performance on this \emph{constructed} benchmark, not on a real-world ontology matching task. The primary evaluation metric for the actual pipeline is \textbf{Precision\,=\,80.19\%}—the fraction of generated relationships verified by a domain expert. Similarity baselines predict ``valid'' when their score exceeds the F1-optimal threshold; the LLM pipeline predicts ``valid'' for all 429 annotated pairs and ``invalid'' for the 500 random pairs.

\begin{table}[ht]
\centering
\caption{Baseline comparison on the balanced evaluation set (929 pairs total). Threshold for similarity methods is set to the F1-optimal value. AUC: Area Under ROC Curve.}
\label{tab:baseline_comparison}
\begin{tabular}{|l|r|r|r|r|}
\hline
\textbf{Method} & \textbf{Prec.} & \textbf{Rec.} & \textbf{F1} & \textbf{AUC} \\ \hline
Jaccard (concept names)     & 0.370 & 1.000 & 0.540 & 0.506 \\ \hline
Jaccard (with context)      & 0.370 & 1.000 & 0.540 & 0.574 \\ \hline
TF-IDF cosine               & 0.576 & 0.552 & 0.564 & 0.625 \\ \hline
DistilBERT (pre-trained)    & 0.582 & 0.515 & 0.546 & 0.609 \\ \hline
Sentence-BERT (MiniLM-L6)   & 0.570 & 0.593 & 0.581 & 0.673 \\ \hline
\textbf{LLM pipeline (ours)}& \textbf{0.802} & \textbf{1.000} & \textbf{0.890} & \textbf{0.927} \\ \hline
\end{tabular}
\end{table}

\begin{figure}[h!tbp]
    \centering
    \includegraphics[width=\linewidth]{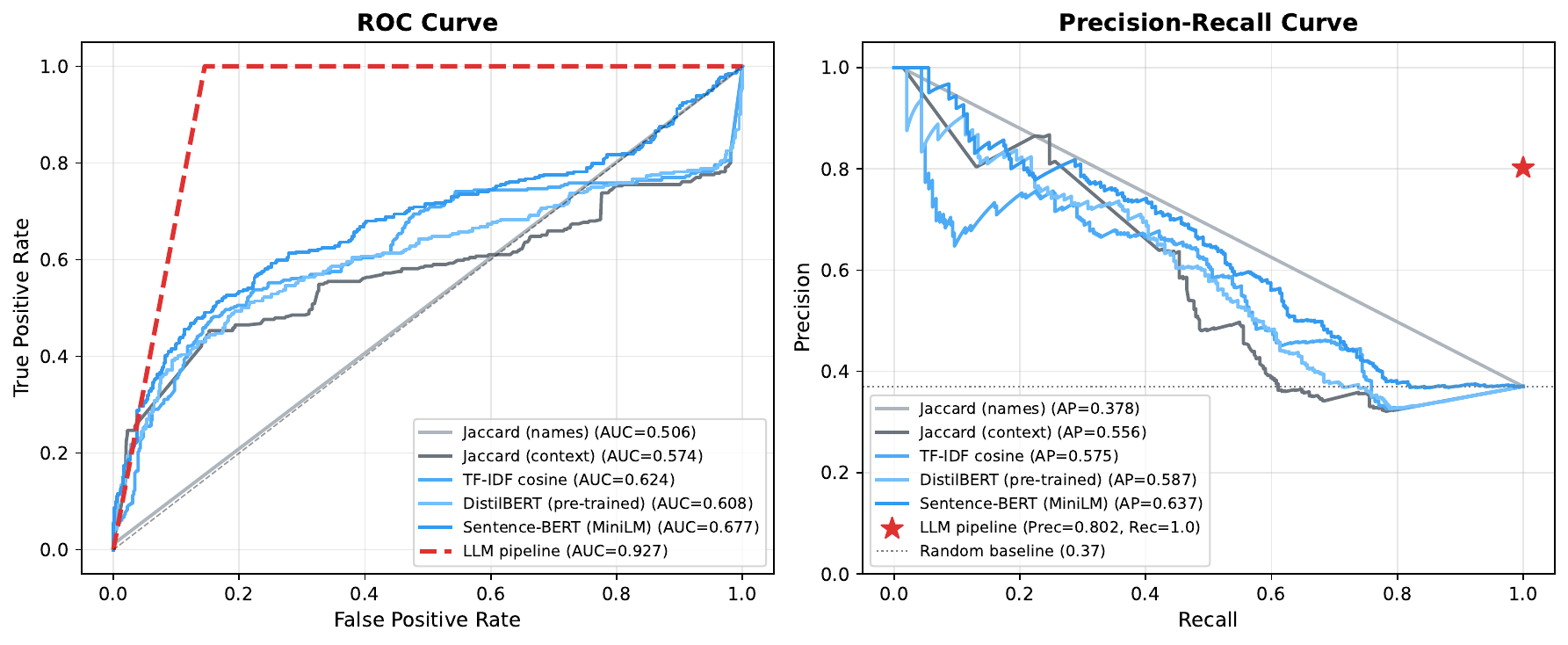}
    \caption{ROC curves (left) and Precision-Recall curves (right) for all baselines and the LLM pipeline on the 929-pair evaluation set.}
    \label{fig:roc_pr}
\end{figure}

The results in Table~\ref{tab:baseline_comparison} show that our LLM pipeline outperforms every similarity-based baseline by a large margin. The best embedding baseline, Sentence-BERT, achieves F1\,=\,0.581; our pipeline achieves F1\,=\,0.890 (+30.9 pp) and Precision\,=\,0.802 (+23.2 pp). The full comparison curves are visualised in Fig.~\ref{fig:roc_pr}.

\subsection{Ablation Study: Why Similarity Alone Is Insufficient}
\label{sec:ablation}

All 429 annotated pairs were pre-selected by the 90th-percentile similarity filter, meaning every pair already has a high cosine similarity score. If similarity were sufficient to judge relationship validity, one would expect an AUC-ROC close to 1.0 when predicting the Validation label from the similarity score on these 429 pairs. Table~\ref{tab:ablation_auc} reports the opposite.

\begin{table}[ht]
\centering
\caption{AUC-ROC for predicting relationship validity from similarity score alone on the 429 pre-filtered pipeline pairs. AUC\,$\approx$\,0.5 confirms similarity cannot distinguish valid from invalid among the high-similarity candidates selected by the pipeline.}
\label{tab:ablation_auc}
\begin{tabular}{|l|r|}
\hline
\textbf{Method} & \textbf{AUC-ROC (on 429 pairs)} \\ \hline
Jaccard (concept names)     & 0.506 \\ \hline
Jaccard (with context)      & 0.485 \\ \hline
TF-IDF cosine               & 0.503 \\ \hline
DistilBERT (pre-trained)    & 0.511 \\ \hline
Sentence-BERT (MiniLM-L6)   & 0.500 \\ \hline
\end{tabular}
\end{table}

\begin{figure}[h!tbp]
    \centering
    \includegraphics[width=\linewidth]{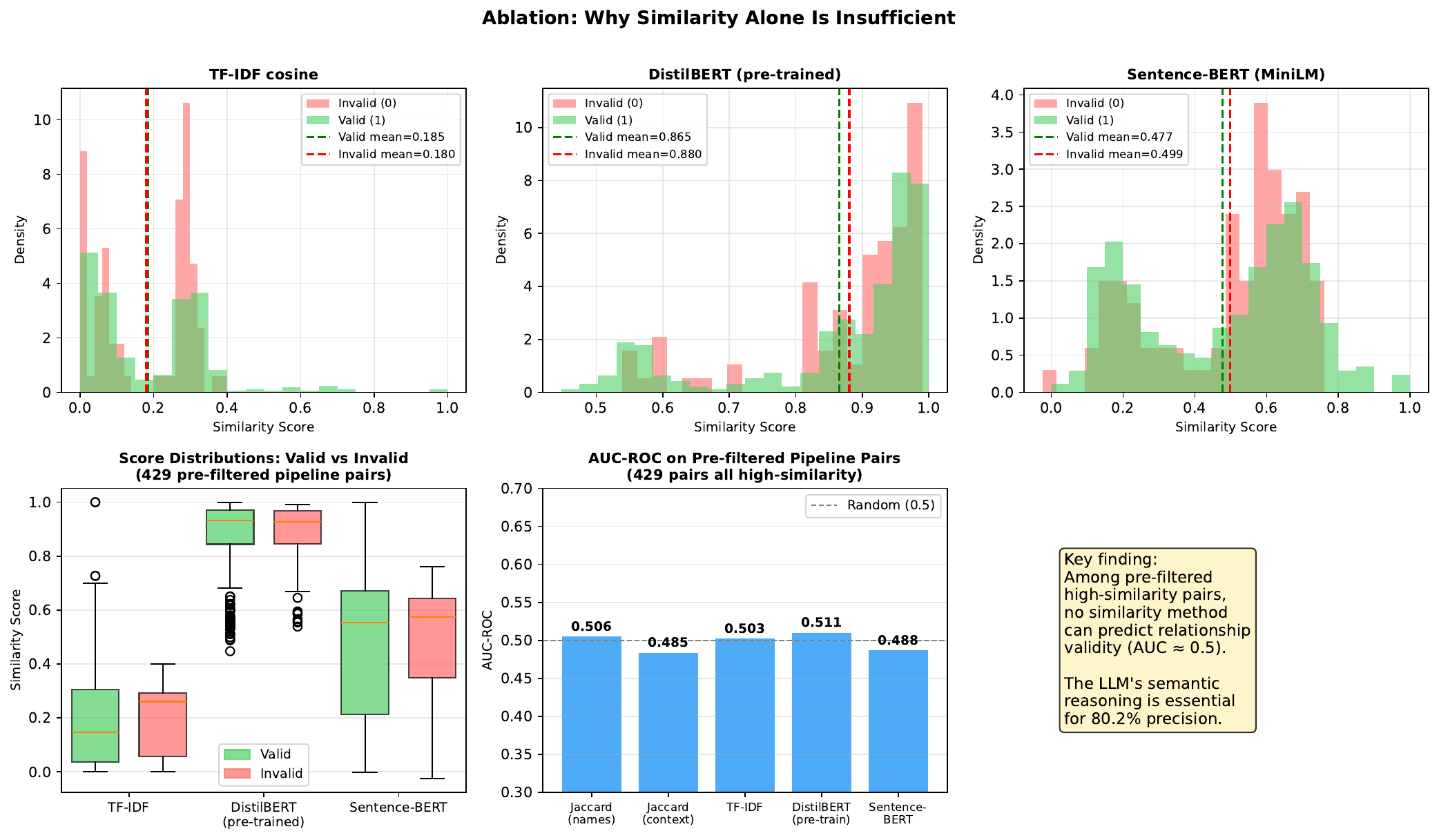}
    \caption{Ablation: similarity score distributions (valid vs invalid, top row), AUC-ROC on pre-filtered pairs (bottom left), and ROC on the full evaluation set (bottom middle). All similarity methods achieve AUC\,$\approx$\,0.5 on pre-filtered pairs, confirming that the LLM step is the critical source of precision.}
    \label{fig:ablation}
\end{figure}

Every method achieves AUC\,$\approx$\,0.5 (random-level), regardless of whether surface-level or deep semantic embeddings are used. This demonstrates that, within the set of high-similarity concept pairs, the signal needed to discriminate valid from invalid relationships \emph{cannot be extracted from similarity scores alone}. Contextual reasoning about concept roles, real-world dependencies, and domain-specific semantics—reasoning that similarity metrics do not provide—is what bridges the gap to 80.2\% precision. Note that the ablation does not rule out graph-based, symbolic, or supervised alternative approaches; it establishes only that similarity scores are insufficient in this setting. Fig.~\ref{fig:ablation} summarises this ablation.

Furthermore, removing the embedding pre-filter makes the pipeline computationally infeasible: the initial cross-domain concept pairing yields approximately 800k candidate pairs. The DistilBERT clustering and 90th-percentile threshold reduce this to $\approx$95k pairs (an 88\% reduction), enabling the LLM step to operate on a focused, high-quality candidate set. The two components are therefore complementary: embeddings provide scalable, semantically grounded candidate selection; GPT-4o contributes typed relationship labelling, reasoning, and justification.

\subsection{Human Annotation and Validation Results}
A sample of 429 generated relationships was independently reviewed by \textbf{two domain expert annotators}—knowledge engineers involved in the ReproduceMeON project with cross-disciplinary expertise in ontology engineering, machine learning, microscopy, and computational science. Each annotator recorded (i) a binary validity judgment ($1 =$ valid, $0 =$ invalid) and (ii) a self-reported certainty score ($1 =$ high certainty, $0 =$ uncertain). Annotator~1 covered all 429 relationships across all four domains. Annotator~2 independently annotated 274 of the 429 relationships across three domains (Machine Learning, Computational, Experimental Workflow); the Microscopy domain (155 pairs) received single-annotator coverage only.

\textbf{Inter-Annotator Agreement (IAA).}
We compute Cohen's $\kappa$ on the 274 dually annotated pairs. Table~\ref{tab:iaa} reports results per domain and overall. The overall Cohen's $\kappa = 0.723$ (90.5\% agreement) falls in the \textit{substantial agreement} range (Landis \& Koch, 1977: $0.61$–$0.80$). The Computational domain achieves $\kappa = 0.817$ (\textit{almost perfect}), while Machine Learning ($\kappa = 0.671$) and Experimental Workflow ($\kappa = 0.720$) both reach the substantial threshold. These results confirm that the annotation task—determining whether a GPT-4o-generated typed relationship is valid—is reliably reproducible by domain experts, and that Annotator~1's judgements are representative.

\begin{table}[ht]
\centering
\caption{Inter-annotator agreement on 274 dually annotated pairs. Cohen's $\kappa$ and percentage agreement per domain and overall. Microscopy (155 pairs) was annotated by a single expert.}
\label{tab:iaa}
\begin{tabular}{|l|r|r|r|}
\hline
\textbf{Domain} & \textbf{n} & \textbf{Cohen's $\kappa$} & \textbf{Agreement (\%)} \\ \hline
Machine Learning     & 137 & 0.671 & 87.6\% \\ \hline
Computational        &  94 & 0.817 & 94.7\% \\ \hline
Experimental Workflow&  43 & 0.720 & 90.7\% \\ \hline
\textbf{Overall}     & \textbf{274} & \textbf{0.723} & \textbf{90.5\%} \\ \hline
\end{tabular}
\end{table}

To supplement the expert assessment, we applied an automated NLI-based validator (BART-large-MNLI~\cite{lewis2019bart}) to all 429 pairs, testing whether each GPT-4o justification textually entails the stated relationship. The validator assigned near-identical entailment scores to valid and invalid pairs (mean NLI score: 0.976 for valid vs.\ 0.970 for invalid), yielding Cohen's $\kappa = 0.0$. This non-result is itself informative—and should be interpreted cautiously: GPT-4o justifications are linguistically persuasive \textit{regardless of correctness}, which constitutes a hallucination risk. A justification that sounds scientifically plausible does not guarantee a valid relationship, and downstream users who rely on justifications alone—without expert review—risk accepting incorrect links. This underscores that domain expertise is irreplaceable for validation and that the interpretability benefit of justifications is weaker than their surface fluency suggests. Overall annotation certainty for Annotator~1 was 87.6\% (376/429 entries rated high certainty).

Since the ontology network has no pre-existing, exhaustive ground truth of all possible cross-ontology links, we report \textbf{Precision} as the primary metric. Recall cannot be computed without a complete gold standard; however, the 429 validated pairs span all four domains and multiple ontology sources, providing indicative evidence of coverage.

\textbf{Overall}: out of 429 evaluated relationships, 344 were validated as correct, yielding an overall precision of \textbf{80.19\%}. When restricting to the 376 high-certainty annotations, precision rises to \textbf{91.49\%}—though this figure should be interpreted with caution, as high-certainty pairs are by definition the less ambiguous cases, making the precision gain expected rather than independently informative.

Table~\ref{tab:precision_results} summarises precision broken down by domain and link type.

\begin{table}[ht]
\centering
\caption{Precision of generated relationships by domain and link type.}
\label{tab:precision_results}
\begin{tabular}{|l|r|r|r|}
\hline
\textbf{Domain} & \textbf{Valid} & \textbf{Total} & \textbf{Precision} \\ \hline
Machine Learning     & 103 & 137 & 75.18\% \\ \hline
Microscopy           & 126 & 155 & 81.29\% \\ \hline
Computational        &  80 &  94 & 85.11\% \\ \hline
Experimental Workflow&  35 &  43 & 81.40\% \\ \hline
\textbf{Overall}     & \textbf{344} & \textbf{429} & \textbf{80.19\%} \\ \hline
\hline
\textbf{Link Type} & & & \\ \hline
Intra-domain         & 192 & 227 & 84.58\% \\ \hline
Inter-domain         & 152 & 202 & 75.25\% \\ \hline
\end{tabular}
\end{table}

Intra-domain precision (84.58\%) is consistently higher than inter-domain precision (75.25\%), which is expected: intra-domain pairs share a common conceptual vocabulary, making semantic alignment clearer. The lower inter-domain precision highlights the greater challenge of linking concepts across heterogeneous ontology spaces and points to a concrete direction for future work, such as domain-adaptive prompting or cross-domain few-shot examples.

\begin{figure}[h!tbp]
    \centering
    \includegraphics[width=\linewidth]{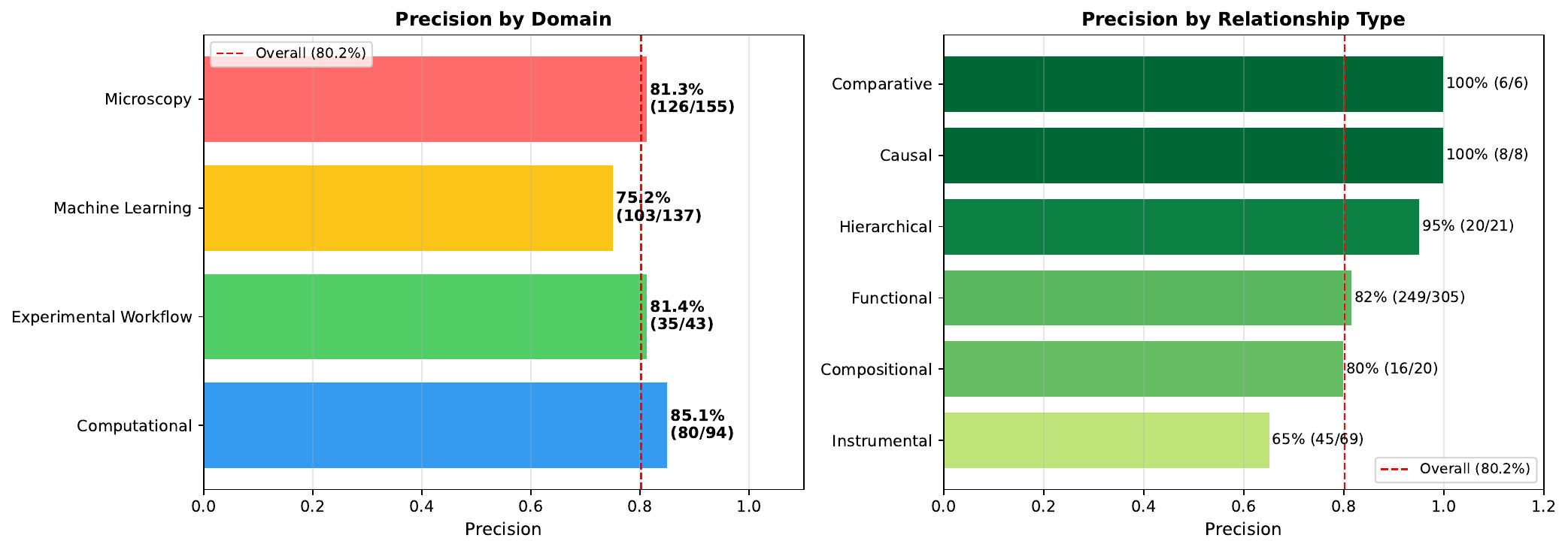}
    \caption{Precision broken down by domain (left) and relationship type (right). Causal, Comparative, and Hierarchical relationships achieve the highest precision; Instrumental relationships are the most error-prone.}
    \label{fig:precision_breakdown}
\end{figure}

Regarding relationship nature, \textit{Causal} (100\%), \textit{Comparative} (100\%), and \textit{Hierarchical} (95.24\%) relationships achieved the highest precision, while \textit{Instrumental} relationships (65.22\%) were the most error-prone—consistent with the inherently ambiguous boundary between tools and processes across domains. Only 19.81\% of the generated links were marked invalid, with a higher error rate in inter-domain contexts, underscoring the need for continued improvement in cross-domain relationship generation. Fig.~\ref{fig:precision_breakdown} summarises precision per domain and per relationship type.

\textbf{Error Analysis.} Examining the 85 invalid relationships, the dominant failure mode was inter-domain linking where similar surface-level terminology masked conceptual incompatibility—most frequently when ML process concepts (e.g., ``Training'', ``Optimization'') were linked to microscopy procedural terms sharing lexical overlap but distinct operational semantics. For Instrumental errors specifically, the model frequently conflated tool-role (an entity \textit{used to} perform a task) with process-role (an entity \textit{that is} a task), particularly across the ML and Experimental Workflow domains where this boundary is ontologically thin. These patterns suggest that domain-adaptive prompting with explicit disambiguation examples for tool vs.\ process concepts could reduce the inter-domain error rate in future work.

\subsection{Limitations}
\label{sec:limitations}

Several limitations should be considered when interpreting our results.
The second annotator covered 274 of 429 relationships; the Microscopy domain (155 pairs) has single-annotator coverage only. Extending dual annotation to Microscopy and computing Microscopy-specific $\kappa$ is planned as future work.
The balanced evaluation set includes 500 randomly generated negatives. Random negatives are easier to reject than real hard negatives (e.g., high-similarity pairs that are genuinely non-relationships), inflating F1. The primary metric—Precision on pipeline outputs—is not subject to this limitation.

All experiments are conducted on ReproduceMeON. While its four-domain, 33-ontology scope is non-trivial, evidence of generalizability to other domains (e.g., biomedical, cultural heritage) or to ontologies of different size and annotation density is lacking. Applying the pipeline to a second domain is planned as future work.
The pipeline depends on GPT-4o, whose stochastic outputs mean that repeating the generation step may yield different relationship labels or justifications for the same concept pair. No repeated-run experiments were conducted to quantify output variance or prompt sensitivity; this affects reproducibility and is a known limitation of LLM-dependent pipelines.

The role taxonomy (Provider, Consumer, etc.) is grounded in established ODPs but its assignment correctness has not been independently evaluated. Competency question testing is identified as future work.

\section{Conclusion}
\label{sec:conclusion}

We presented an end-to-end pipeline for enhancing ontology network development by automatically discovering intra- and inter-domain relationships. The pipeline combines fine-tuned DistilBERT embeddings, clustering-based candidate selection, iterative GPT-4o prompt engineering, and role-based relationship materialization. Applied to ReproduceMeON across four domains (computational, machine learning, microscopy, and experimental workflow), dual expert validation of 429 generated relationships confirms an overall precision of 80.19\% (rising to 91.49\% on high-certainty annotations), with an F1 of 0.890 on a balanced evaluation set. Inter-annotator agreement on 274 dually annotated pairs reaches Cohen's $\kappa = 0.723$ (\textit{substantial agreement}), confirming that the annotation task is reliably reproducible. A systematic quantitative comparison against five similarity-based baselines—including Sentence-BERT and pre-trained DistilBERT—confirms that the best embedding baseline achieves F1\,=\,0.581, demonstrating that our LLM pipeline outperforms all similarity-based alternatives by over 30 percentage points in F1. An ablation study further shows that among pre-filtered, high-similarity concept pairs, all similarity methods achieve AUC\,$\approx$\,0.5, confirming that semantic similarity alone is insufficient to judge relationship validity: the LLM's reasoning about concept roles and domain dependencies is what delivers the precision gain. The embedding pre-filtering step remains essential for computational feasibility, reducing 800k raw candidate pairs to 95k high-confidence pairs. Future work will explore alternative LLMs (e.g., open-source instruction-tuned models) for cost-effective deployment, full recall measurement via a curated gold-standard dataset, integration of provenance domain ontologies, and extension of role-based mapping to support automated workflow generation and SPARQL-based reasoning over the enriched ontology network.

\textbf{Declaration on Generative AI:} The text of this manuscript was improved with the following AI tools: ChatGPT and Claude. After using these tools, the authors reviewed and edited the content as needed and take full responsibility for the publication’s content.

\bibliography{literature}

\end{document}